\documentclass[10pt,twocolumn,letterpaper]{article}

\usepackage{cvpr}
\usepackage{times}
\usepackage{epsfig}
\usepackage{graphicx}
\usepackage{amsmath}
\usepackage{amssymb}
\usepackage{subfig}
\usepackage{fixmath}
\usepackage{color}
\usepackage[keeplastbox]{flushend}

\usepackage[pagebackref=true,breaklinks=true,letterpaper=true,colorlinks,bookmarks=false]{hyperref}

\cvprfinalcopy 

\newcommand{\Lagr}{\mathcal{L}}

\def\cvprPaperID{2584} 

\ifcvprfinal\fi

\begin{document}

\title{One-shot Face Recognition by Promoting Underrepresented Classes
}

\author{Yandong Guo, Lei Zhang\\
Microsoft\\
One Microsoft Way, Redmond, Washington, United States\\
{\tt\small \{yandong.guo, leizhang\}@microsoft.com}
}

\maketitle

\begin{abstract}

In this paper, 
we study the problem 
of training 
large-scale face identification model 
with imbalanced training data. 
 This problem naturally exists 
in many real scenarios 
including large-scale celebrity recognition, movie actor annotation, etc. 
Our solution contains two components. 
First, we 
build a face feature extraction model, 
and improve its performance, 
especially for the persons with very limited training samples, 
by introducing a regularizer to the cross entropy loss  for the multinomial logistic regression (MLR) learning. 
This regularizer encourages the directions of the face features from the same class to be close to the direction of their corresponding classification weight vector in the logistic regression. 
Second, we build a multiclass classifier using
MLR on top of the learned face feature extraction model. 
Since the standard MLR has poor generalization capability for the one-shot classes even if these classes have been oversampled, 
we propose a novel supervision signal 
called underrepresented-classes promotion loss,
which aligns the norms of the weight vectors of the 
one-shot classes (a.k.a. underrepresented-classes)
to those of the normal classes. 
In addition to the original cross entropy loss, 
this new loss term
effectively promotes the underrepresented classes in the learned model and leads to a remarkable improvement in face recognition performance. 

We test our solution on 
the MS-Celeb-1M low-shot learning benchmark task.
Our solution 
recognizes $94.89\%$ of the test images
at the precision of $99\%$ 
for the one-shot classes. 
To the best of our knowledge, this is 
the \textbf{best} performance
among all the published methods 
using this benchmark task
with the same setup, 
including all the participants in the recent MS-Celeb-1M challenge at ICCV 2017. 

\end{abstract}


\section{Introduction}
\label{sec:intro}

The great progress of face recognition in recent years has made large-scale face identification possible for many practical applications. In this paper, we study the problem of
training a large-scale face identification model
using \textit{imbalanced} training images for a large quantity of persons, 
and then use this model to identify other face images
for the persons in \textit{the same group}. 
This setup is widely used 
when the images for the persons to be recognized are available beforehand, 
and an accurate recognizer is needed for a large
and relatively fixed group of persons. 
For example, 
large-scale celebrity recognition for search engine, 
public figure recognition for media industry, 
and movie character annotation for video streaming companies. 

Building a large-scale face recognizer is 
not a trivial effort.
One of the major challenges is caused by 
the highly imbalanced training data. 
When there are many persons to be recognized, 
it naturally happens that 
for some of the persons to be recognized, 
there might be very limited number of training samples,
or even only one sample
for each of them.  
Besides this unique challenge, 
there are also other 
challenges
introduced by  
the fact that different persons may have very similar faces, 
and the fact that the faces from the same person 
may look very different due to lightning, pose, and age variations. 

To study this problem, we design a benchmark task
and propose a strong baseline solution for this task. Our benchmark task 
is to train a face recognizer to identify $21,000$ persons. 
For the $20,000$ persons among them, 
we provide
about $50$-$100$ training images per person
and call this group \textit{base set}, following the terminology defined in \cite{Ross2016lowshot}. 
For the other $1,000$ persons, 
we only offer \textbf{one} training image per person, 
and call this group \textit{low-shot set}. 
The task is to study 
with these training images only, 
how to develop an algorithm to recognize the persons in \textbf{both} data sets. 
In particular, we mainly focus on the recognition accuracy for persons in the low-shot set as it shows the one-shot learning capability of a vision system, while we also check the recognition accuracy for those in the base set to ensure not to hurt their performance.
We have published this data set to facilitate the research in this direction. 

Our solution for this benchmark task 
is to train a good face representation model
and build a classifier on top of that. 
The objective of feature learning 
is to train a face representation model 
with good discrimination ability  
not only for the base set, 
but also for the low-shot set. 
In other words, 
since there is only one training image per person 
in the low-shot set, 
we need to build face feature extractor 
with good generalization capability. 
There have been a lot of effort in this direction,
yet our method is different in the following two perspectives. 
First is data.
We 
train our face feature extractor with 
the base set, 
which includes
about one million images 
with high annotation accuracy.  
This is one of the \textit{largest public available} datasets
\cite{LFWTech,LFWTechUpdate,Youtube,CASIA_WebFace,Yandong:Celeb,Xiaoou_Deep1}, 
which makes our model reproducible and meaningful. 
Second is the cost function design. 
In addition to the standard cross entropy loss used together with Softmax for multinomial logistic regression (MLR) learning, 
we propose to add another loss term, 
which 
encourages features of the same classes to have similar directions
to their corresponding weight vector 
in logistic regression.
Since the weight vector is trained to 
have the direction
which is close to the direction of the features 
from its corresponding class, 
and 
far away from the directions of the features from other classes, 
our proposed term effectively minimizes the intra-class variance and maximizes the inter-class variance simultaneously. 
We compare our face representation model 
with its most similar alternative methods 
and demonstrate the advantages of our method in subsection 
\ref{sec:related-feature}, 
\ref{sec:method-feature}, 
and \ref{sec:experiments-feature}.


The second stage of our solution is to learn
a classifier on top of the face feature extractor learned in the first stage. 
Though K-nearest neighborhood (KNN)
or other template-based methods 
might be the most straight-forward solution, 
the standard KNN method is not suitable for 
our setup 
due to its limitations  
in aspects of accuracy and scalability \cite{ACMMMMSCeleb1M-2,ICCV-W-Yue,Xu_2017_ICCV}. 
More discussions are presented in section 
\ref{sec:relatedKNN}. 
In our solution, we choose to use 
MLR for its proven great performance on various visual recognition problems. 

The major challenge of using MLR as the classifier 
is caused 
by the highly imbalanced training data.  
In our experiments, we have observed
almost perfect performance of MLR in recognizing persons 
in the base set, 
yet
very poor performance of MLR for the low-shot set, 
even though their training images are oversampled. 
A further analysis in Section \ref{sec:methodology} shows that a low-shot class with only one training sample can only claim 
small partition in the feature space. Moreover, we reveal that there is a close connection between the volume of a class partition in the feature space and the norm of the weight vector of this class in the 
MLR model. Based on this finding, we propose to add a new loss term to the original cross-entropy loss for MLR, serving as the prior for the weight vectors in multinomial logistic regression. This new loss term is based on our empirical assumption and observation that on average, each person in the low-shot set should occupy a space of similar volume in the feature space, compared with the persons in the base set. We call this term the Underrepresented-classes Promotion (UP) loss. 
For comparison, we also explore other different options for the priors of the weight vectors. 

To quantitatively evaluate the performance, we adopt the close-domain face identification
setup, 
and apply the classifiers 
with the test images \textit{mixed} 
from both the base set ($100,000$ images, $5$ images/person) 
and the low-shot set ($20,000$ images, $20$ images/person). 
Our experimental results clearly demonstrate the effectiveness of the proposed method. 
With our feature extraction model and the UP term, we can recognize 
$94.89 \%$
of the test images in the low-shot set
with a high precision of $99\%$
and keep the top-1 accuracy of $99.8\%$ for the base classes, 
while 
without using our method, 
only $25.65\%$ of the test images from the low-shot set 
can be recognized at the same precision.


In summary, our contributions can be highlighted as follows. 
\begin{itemize}
\item We set up a benchmark task for one-shot face recognition, and provide the associated dataset
composed of the base set and the low-shot set. 

\item 
We propose a new cost function 
to effectively 
learn discriminative feature extractor
with good generalization capability on the low-shot set.

\item We reveal that the deficiency of the multinomial logistic regression (MLR) in one-shot learning is related to the norms of the weight vectors in 
MLR, and propose a novel loss term called underrepresented-classes promotion (UP) which effectively addresses the data imbalance problem in one-shot learning.

\item 
Our solution 
recognizes $94.89\%$ of the test images
at the precision of $99\%$ 
for the low-shot classes. 
To the best of our knowledge, this is 
the \textbf{best} performance
among all the published methods 
using this benchmark task with the same setup. 

\end{itemize}

\section{Related Work}
\label{sec:related}
\subsection{Benchmark Task}

Nowadays, we observe 
the major focus in face recognition 
has been to learn a good face feature extractor.
In this setup, typically, 
a face feature extractor is trained with images for a group of persons, 
and then tested with images for a \textit{different} group of persons in the verification or identification task. 
For example, 
the verification task with the LFW dataset \cite{LFWTech}
is the de facto standard test to evaluate face features, though the performance on this dataset
is 
getting saturated.   
Moreover, a lot of face identification tasks, 
e.g., MegaFace \cite{UW_MegaFace} or LFW \cite{LFWTech} with the identification setup,
are essentially to evaluate face features
since the identification is achieved by 
comparing face features between query and gallery images. 

The major advantage of the above setup 
is that the generalization capability of face representation model
can be clearly evaluated, since the persons in the training phase are usually \textit{different} from the persons in the testing phase. This is very important 
when the images of the target persons are not accessible during the training phase.

Unfortunately, 
we observe the best performance for the
above setup
is typically obtained by using very large, \textit{private} dataset(s), which makes it impossible to reproduce these work, e.g., \cite{Google_Face}. 
Moreover, though to obtain a good feature extractor is essential and critical for face identification, 
good feature extractor is not yet the final solution for the identification. 

Our benchmark task has a different setup. 
we train the face identification model with 
the imbalanced training images for the persons to be recognized. 
This setup is very useful when the images for the target persons are available beforehand,
because it generally leads to better performance to train with images for the target persons
compared with to train with images 
for other persons (assuming similar total amount of images).
As discussed in the introduction section, there are also many real scenarios using this setup. 
Moreover, since in our task we include the
low-shot classes (persons with only one training sample), the generalization capability 
can also be evaluated. 
Last but not least, we provide both the training and testing datasets, so people can conveniently reproduce and compare their algorithms in this direction. 

\subsubsection{Low-shot learning for general visual recognition}
In the general image recognition domain, 
the recent low-shot learning work \cite{Ross2016lowshot} also attracts a lot of attentions. 
Their benchmark task is very similar to ours but in the general image recognition domain:
the authors split the ImageNet data \cite{ILSVRC15} into the base and low-shot (called novel in \cite{Ross2016lowshot}) classes, 
and the target is to recognize images from both the base and low-shot classes. 
Their solution is quite different from ours since the domain is quite different. 
We will not review their solution here due to the space constraint, but list results from their solution as one of the comparisons in the experiments section \ref{sec:experiments}.

\subsection{Discriminative Feature learning}
\label{sec:related-feature}
Cross entropy with Softmax has demonstrated good performance in supervising the face feature extraction model training. 
In order to further improve the performance of representation learning, 
many methods have been proposed
to add extra loss terms or slightly modify the cross entropy loss (used together with softmax for multinomial logistic regression learning) to regularize the representation learning in order to improve the feature discrimination and generalization capability.

Among all these works, we consider 
the center loss \cite{centerECCV} as one of the 
representative methods (a similar idea published in \cite{Latha:dense} during the same time).
In \cite{centerECCV}, face features from the same class are encouraged to be close 
to their corresponding class center (actually, approximation of the class center, usually dynamically updated). 
By adding this loss term to the standard Softmax, the authors obtain a better face feature representation model \cite{centerECCV}. 

There are many other alternative methods, including 
the range loss in \cite{RangeLoss}, 
fisher face in \cite{Fisher}, 
center invariant loss in \cite{centerinvariant}, 
marginal loss in \cite{MarginalLoss}, 
sphere face in \cite{sphereface}, 
etc.
Each of these methods has its own uniqueness and advantages under certain setup. 

We design a different kind of loss term adding to the cross entropy loss of the Softmax 
to improve the feature extraction performance. 
In section \ref{sec:methodology} 
and \ref{sec:experiments}, 
we demonstrate that our method 
has better performance in our setup 
than 
that with 
center loss in \cite{centerECCV} or sphere face in \cite{sphereface}
(these two are the most similar ones)
from the perspective of theoretical discussion and experimental verification. 
Our method has only one parameter and is very easy to use. 
We have not reproduced all these 
cost function design methods \cite{RangeLoss,Fisher,centerinvariant,MarginalLoss} with our training dataset due to practical reasons. 
These methods were implemented 
with different networks structures, 
and trained on different datasets. 
Sometimes parameter adjustment is critically required when the training data is switched.
We will work on evaluating more methods in the future. 

\subsection{KNN vs. Softmax}
\label{sec:relatedKNN}
After a good face feature extractor is obtained, 
the template-based method, 
e.g., K-nearest neighborhood (KNN), 
is widely used for face identification these days.
The advantages of KNN is clear:
no classifier training is needed, 
and KNN does not suffer much from imbalanced data, etc.
However, experiments in \cite{ICCV-W-Yue,ACMMMMSCeleb1M-1,ACMMMMSCeleb1M-2,Xu_2017_ICCV} and section \ref{sec:experiments} in our paper
demonstrate that the accuracy of KNN
with the large-scale face identification setup
is usually lower than MLR, when the same feature extractor is used. 
Moreover, 
if we use all the face images for every person in the gallery, 
the complexity is usually too high for large scale recognition,
and the gallery dataset needs to be very clean to ensure high precision. 
If we do not keep all the images per person, 
how to construct representer for each class is still an open problem. 

As described above, 
MLR demonstrates overall higher accuracy compared with KNN
in many previous publications. 
This is mainly because 
in MLR, the weight vectors for each of the classes
is estimated using discriminant information from all the classes, 
while in the KNN setup, 
the query image only needs to be close enough to one local class 
to be recognized. 
Moreover, after feature extraction, 
with MLR, 
the computational complexity of estimating the persons' identity 
is linear to the number of persons, 
not the number of images in the gallery. 

However, the standard MLR classifier 
suffers from the imbalanced training data
and has poor performance 
with the low-shot classes
even these classes are oversampled during training, 
though the overall accuracy is higher than KNN. 
Recently, some works develop hybrid solutions by combining MLR and KNN \cite{ICCV-W-Yue,Xu_2017_ICCV} and achieve promising results. 
In these work, when MLR does not have high confidence (threshold tuning is needed), 
KNN is used. 

We solve this problem from a different perspective.
Different from the hybrid solution, our solution only has one MLR as the classifier so that no threshold is needed to switch between classifiers. 
We boost the performance of MLR by regularizing the norm of the weight vectors in MLR. We have not seen a lot of effort in this direction, especially in the deep learning scenario. 




\section{Benchmark Datasets}
We have prepared and published the associated datasets for our task
earlier this year, 
and attracted more than one hundred downloads. Here we clarify the key points of our dataset for everyone's convenience. 

\noindent 
\textbf{Training}

There are $21$K persons in total. 
For the $20$K persons among them, 
there are $50$-$100$ training images per person (base set). 
For the rest $1000$ persons, 
there is one training image per person (low-shot set). 

\noindent
\textbf{Testing}

We test the face identification with the same $21$K persons. 
There are $120$K images to be recognized ($100$K from the base set and $20$K from the low-shot set). 
The model to be tested will not have access to know 
whether the test image is from the base set or the low-shot set, which is close to the real scenario, 
yet the performance is evaluated on base and low-shot separately
to better understand the system. 

\noindent
\textbf{Comparison}

Though the base set in this paper is considerably larger than most of the public face dataset, 
it is smaller than MS-Celeb-1M \cite{guo2016msceleb} (actually a subset of MS-Celeb-1M). 
The base set has a different focus from MS-Celeb-1M. 
MS-Celeb-1M targets at recognizing as many as possible celebrities in the one-million celebrity list so the celebrity coverage is important, 
and the noisy label for the less popular celebrity is inevitable \cite{ACMMMMSCeleb1M-1,ACMMMMSCeleb1M-2,MSCelebNoiseLabel}.
Therefore, MS-Celeb-1M inspires work including data cleaning, training with noisy-labels, etc. 
On the other hand, 
the base set published in this paper 
is mainly for training a robust and generalizable face feature extractor, as it is nearly noise-free.

Moreover, for the convenience of feature evaluation, we do not include the celebrities in LFW \cite{LFWTech} (the de facto standard) in our 
base set ($20$K persons).
Thus researchers can directly leverage this dataset and evaluate performance on the LFW verification task. 

Our low-shot set can also be used to evaluate feature extractors (though indirectly) since only one image is provided per person in the low-shot set. We provide $20$ images per person in this test for testing purpose. 
For comparison, the LFW dataset \cite{LFWTech},
has less than $100$ persons having more than $20$ images.
The benchmark task in MegaFace \cite{UW_MegaFace} focuses on $80$ identities for the query set to be recognized, though millions of images provided as distractors.

\section{Methodology}
\label{sec:methodology}
Our solution includes the following two phases. 
The first phase is \textit{representation learning}. 
In this phase, we build face representation model using all the training images from the \textit{base set}. 

The second phase is \textit{one-shot learning}. 
In this phase, 
we train a multi-class classifier, 
with the help our UP term,
to recognize the persons in both \textit{base set} 
and \textit{low-shot set} 
based on the representation model learned in phase one.

\subsection{Representation learning}
\label{sec:method-feature}
We train our face representation model 
with supervised learning framework considering persons' ids as class labels. 
The cost function we use is 
\begin{equation}
\label{eq:costfunction}
\Lagr =  
\Lagr_s + \lambda \Lagr_a
 \, ,  
\end{equation}
where $\Lagr_s$ is the standard cross entropy loss used for the Softmax layer, 
while $\Lagr_a$ is our proposed loss used to 
improve the feature discrimination and generalization capability, with the balancing coefficient $\lambda$. 

More specifically, we recap the first term, cross entropy $\Lagr_s$
as 
\begin{align}
\label{eq:crossentropy}
\Lagr_s &= - \sum_n \sum_k t_{k,n} \log p_{k}(x_n) \,,
\end{align}
where $t_{k,n}\in\{0,1\}$ is the ground truth label indicating whether the $n^{th}$
image belongs to the $k^{th}$ class, 
and the term $p_{k}(x_n)$
is 
the estimated probability that the image $x_n$
belongs to the $k^{th}$ class,
defined as, 
\begin{equation}
\label{eq:sigmoid}
p_{k}(x_n) = \frac{\exp (\mathbf{{w}}_k^T {\boldsymbol{\phi}(x_n)} )}{\sum_i \exp (\mathbf{w}_k^T {\boldsymbol{\phi}(x_n)})} \, ,
\end{equation}
where 
$\mathbf{w}_k$ is the weight vector for the $k^{th}$ class, 
and
$\boldsymbol{\phi}(\cdot)$ denotes the feature extractor for  image $x_n$. 
We choose the 
standard residual network with 34 layers (ResNet-34) \cite{Resnet} as our feature extractor 
$\boldsymbol{\phi}(\cdot)$
using the last pooling layer as the face representation.
ResNet-34 is used due to its good trade-off between prediction accuracy and model complexity, yet our method is general enough to be extended to deeper network structures for even better performance.
Note that in all of our experiments, we always set the bias term $b_k=0$. 
We have conducted comprehensive experiments 
and found
that removing the bias term 
from the standard Softmax layer 
does not affect the performance. 
yet leads to a much better understanding of the geometry property of the classification space.

The second term $\Lagr_a$
in the cost function \ref{eq:costfunction}
is calculated as
\begin{align}
\label{eq:cca}
\mathbf{w}'_k &\leftarrow \mathbf{w}_k \\
\Lagr_a &= - \sum_k\sum_{i \in C_k} \frac{\mathbf{w}_k^{'T} \boldsymbol{\phi}(x_i)}{\|\mathbf{w}'\|_2 \|\boldsymbol{\phi}(x_i)\|_2} \, .
\end{align}
We set the parameter vector $\mathbf{w}'_k$ to be equal to the weight vector $\mathbf{w}_k$.
This loss term encourages the face features belong to the same class to have similar direction 
as their associated classification weight vector $\mathbf{w}_k^T$. 
We call this 
term as 
Classification vector-centered Cosine Similarity (CCS) loss.
Calculate the derivative with respect to $\boldsymbol{\phi}(x_i)$, we have 
\begin{equation}
\frac{\partial \Lagr_a}{\partial \boldsymbol{\phi}(x_i)}=\frac{1}{\|\boldsymbol{\phi}(x_i) \|_2} 
\left(
\frac{\mathbf{w}_k^{'T}}{\|\mathbf{w}'_k\|_2}  
-\frac{\boldsymbol{\phi}(x_i)^T \cos \theta_{i,k}}{\|\boldsymbol{\phi}(x_i)\|_2} 
\right) \,,
\end{equation}
where $\theta_{i,k}$ is the angle between $\mathbf{w}'_k$
and $\boldsymbol{\phi}(x_i)$. 
Note that $\mathbf{w}'_k$ in this term is the parameter copied  from $\mathbf{w}_k$, so there is no derivative to $\mathbf{w}'_k$. 
For experiment ablation purpose, we also tried to back propagate the derivative of $\mathbf{w}_k$, but did not observe better results.

\subsubsection{Discussion}
There have been a lot of effort in 
adding extra terms 
to cross entropy loss
to improve the feature generalization capability. 
Maybe the most similar version 
is the center loss in \cite{centerECCV}, 
also known as 
the dense loss in 
\cite{Latha:dense} published during the same time.
In center loss, the extra term is defined as 
\begin{equation}
\label{eq:center}
\Lagr_c = - \sum_k\sum_{i \in C_k} ||\mathbf{c}_k - \boldsymbol{\phi}(x_i) ||_2^2\, ,
\end{equation}
where $\mathbf{c}_k$ is defined as the \textit{class} center (might be dynamically updated as the approximation of the true class center due to implementation cost).

Our method is different from center loss from two perspectives. 
First, minimizing the cost function \ref{eq:center}
may lead to two consequences. While it helps reduce the distance between
$\boldsymbol{\phi}(x_i)$ and its associated center $\mathbf{c}_k$, it also reduces the norms 
of $\boldsymbol{\phi}(x_i)$ and $\mathbf{c}_k$. 
The second consequence is usually not good as it may hurt the classification performance. 
We did observe in our experiment that 
over training with center loss 
would lead to features with too small norms 
and worse performance compared with not using center loss (also reported in \cite{centerECCV}). 
On the contrary, our loss term only considers the angular between 
$\boldsymbol{\phi}(x_i)$ and $\mathbf{w}'_k$, and will not affect the norm of the feature. 
In our experiment section, we demonstrate that our method is not sensitive to the parameter tuning. 

Second, 
please note that we use the weight vector in Softmax
$\mathbf{w}_k$ to represent the \textbf{classification} center, while in \ref{eq:center}, 
the variable $\mathbf{c}_k$
is the \textbf{class} center. 
The major difference is that 
$\mathbf{w}_k$ is updated 
(naturally happens during minimizing 
$\Lagr_s$) 
using not only the information from the $k^{th}$ class, but also the information from the other classes.
In contrast, 
$\mathbf{c}_k$ is updated only using the information from the $k^{th}$ class (calculated separately). 
More specifically, 
according to the derivative of the cross entropy loss in \ref{eq:crossentropy}, 
\begin{equation}
\label{eq:gcd}
\frac{\partial \Lagr_s}{\partial \mathbf{w}_k}=\sum_n (p_{k}(x_n)-t_{k,n}){\boldsymbol{\phi}(x_n)}  \,, 
\end{equation}
the direction of $\mathbf{w}_k$ 
is close to the direction of the face features from the $k^{th}$ class, and being pushed far away from the directions 
of the face features \textit{not} from the $k^{th}$ class.

\subsection{One-shot Learning}
In this subsection, we build a classifier
using multinomial logistic regression on top of the feature representation model we obtained in the previous subsection.

\subsubsection{Challenges of One-shot}


As we discussed previously, 
the standard MLR
does not perform 
well for the persons in the low-shot set. 
In section \ref{sec:experiments}, 
we report that 
with the standard MLR, 
for the low-shot set, 
the coverage at the precision of $99\%$
is only $25.65\%$, 
while for the base set, 
the coverage is $100\%$ at the 
precision of $99\%$. 
Note that the training images in the low-shot set have been oversampled by $100$ times. 
The feature extractor with standard softmax was used.


The low coverage for the low-shot classes 
is related to the
fact that the only one sample from each low-shot class occupies a much smaller partition in the feature space, compared with the samples in each base class. This is because a class with one sample usually has a much smaller (even 0 for one sample) intra class variance than a classes with many samples which can span a  larger area in the feature space. 
To further understand this property, 
without loss of generality, we
discuss the decision hyperplane between any two adjacent classes. 
We apply Eq. \ref{eq:sigmoid} to both the $k^{th}$ class and the $j^{th}$ class to determine the decision hyperplane between the two classes (note we do not have bias terms throughout our paper):
\begin{equation}
\label{eq:decision}
\frac{p_j(x)}{p_k(x)}=\frac{\exp (\mathbf{{w}}_j^T {\boldsymbol{\phi}(x)} )}{\exp (\mathbf{{w}}_k^T {\boldsymbol{\phi}(x)} )}=\exp [(\mathbf{{w}}_j - \mathbf{{w}}_k)^T
\boldsymbol{\phi}(x)]
\end{equation}

\begin{figure}
\centering
\vspace*{-0.36in}
\subfloat[$\|\mathbf{w}_k\|_2 =\|\mathbf{w}_j\|_2 $]{\includegraphics[width=0.45\linewidth]{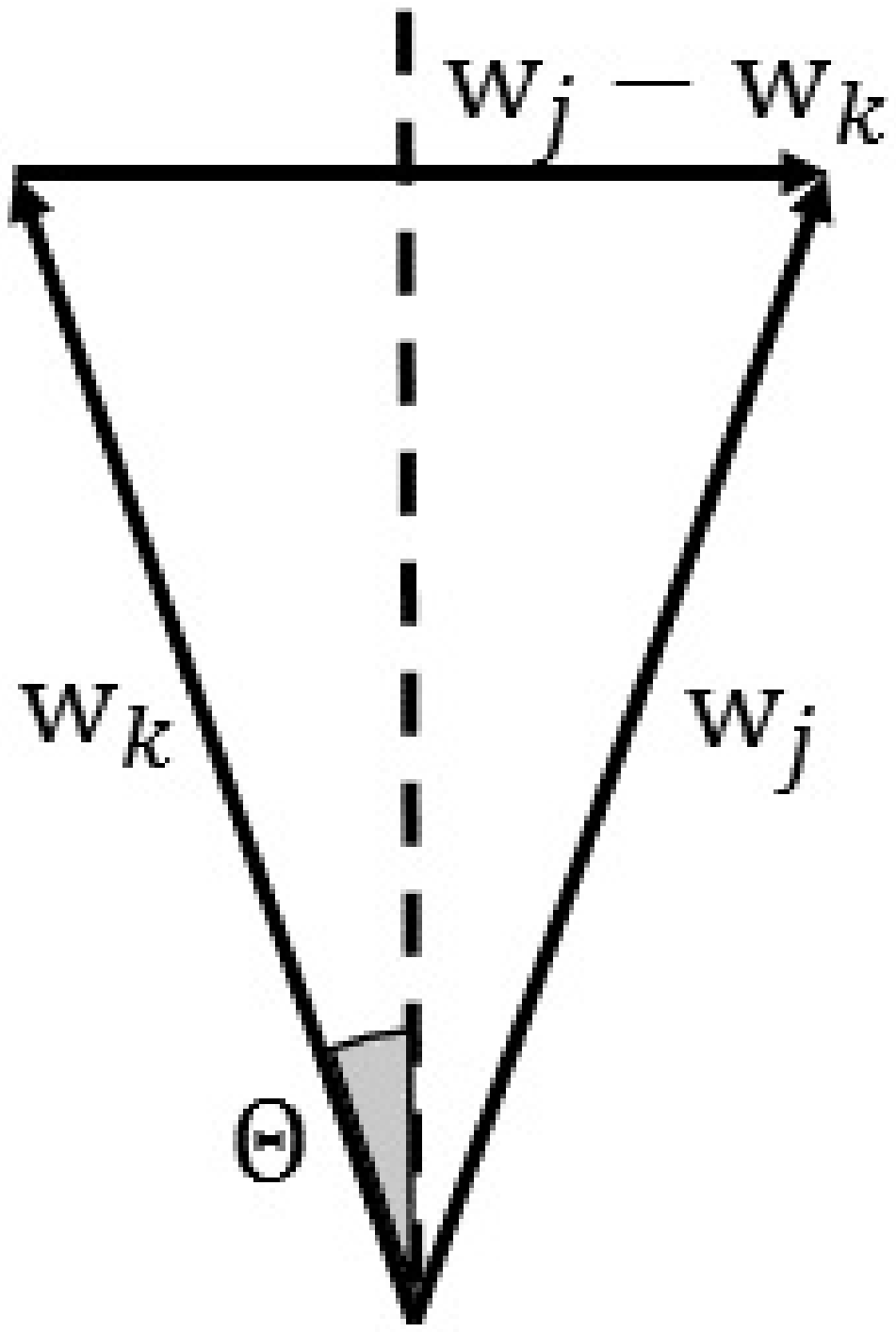}} \hspace{-1cm}
\subfloat[$\|\mathbf{w}_k\|_2 <\|\mathbf{w}_j\|_2 $]{\includegraphics[width=0.45\linewidth]{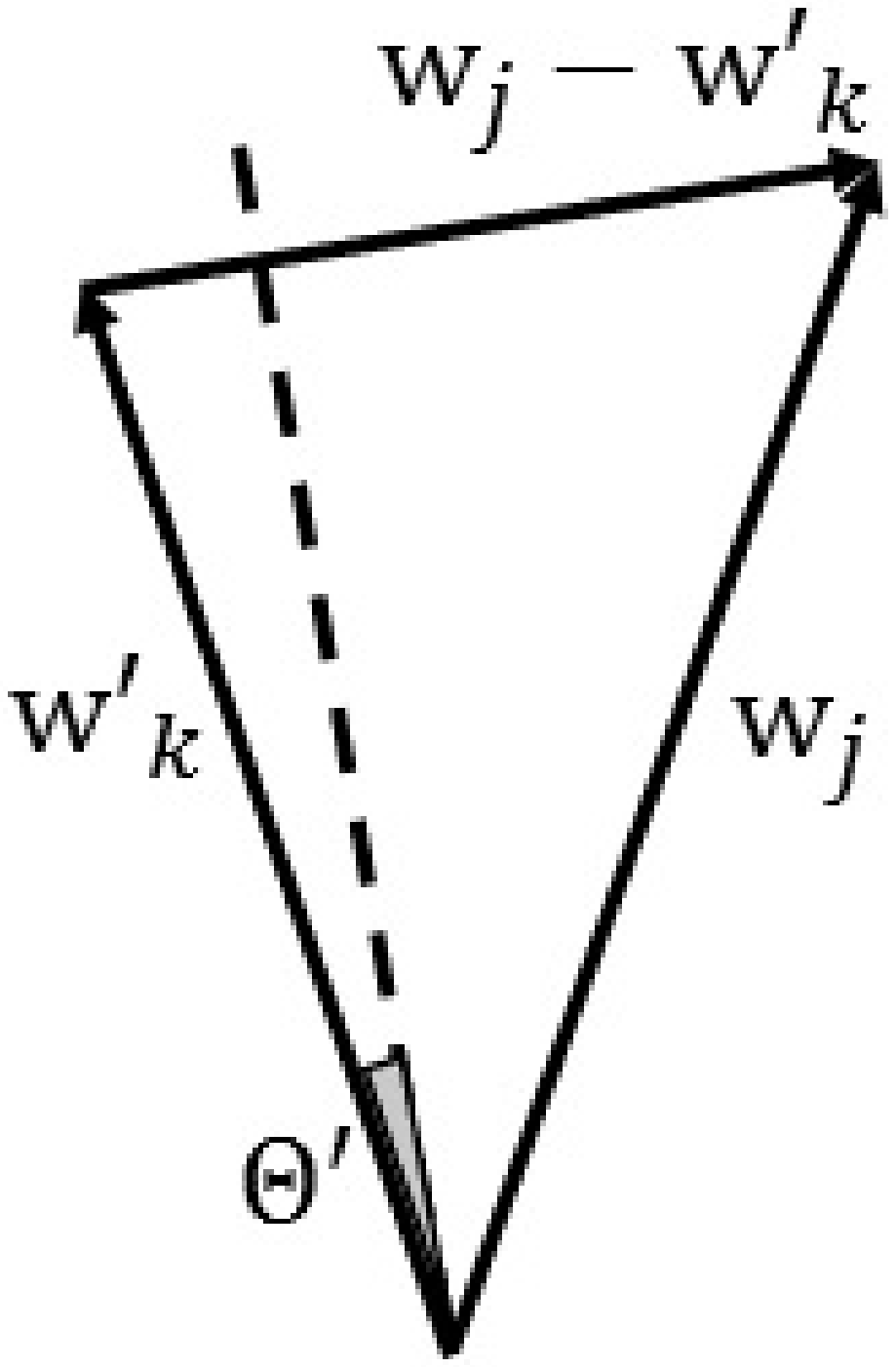}}
\caption{Relationship between the norm of $\mathbf{w}_k$
and the volume size of the partition for the $k^{th}$ class.
The dash line represents the hyper-plane (perpendicular to $\mathbf{w}_j-\mathbf{w}_k$) which separates the two adjacent classes. 
As shown, 
when the norm of $\mathbf{w}_k$ decreases, 
the $k^{th}$ class tends to possess a smaller volume size
in the feature space. 
}
\label{fig:weight}
\end{figure}

As shown in Figure \ref{fig:weight}, the hyperplane
to separate two adjacent classes $k$ and $j$
is perpendicular to the vector $\mathbf{w}_j - \mathbf{w}_k$. 
When the norm of $\mathbf{w}_k$ gets decreased, 
this hyperplane is pushed towards the $k^{th}$ class, 
and the volume for the $k^{th}$ class 
gets decreased. 
As this property holds for any two classes, we can clearly see the connection of the norm of a weight vector and the volume size of its corresponding partition space in the feature space.

In our experiments with the standard MLR, 
we found that 
the norms 
of the weight  
vectors for the low-shot classes are 
much 
smaller than the norms of the weight vectors
for the base classes,
with an example shown in Figure \ref{fig:wnorm1}.

\begin{figure}
\centering
{\includegraphics[width=0.8\linewidth]{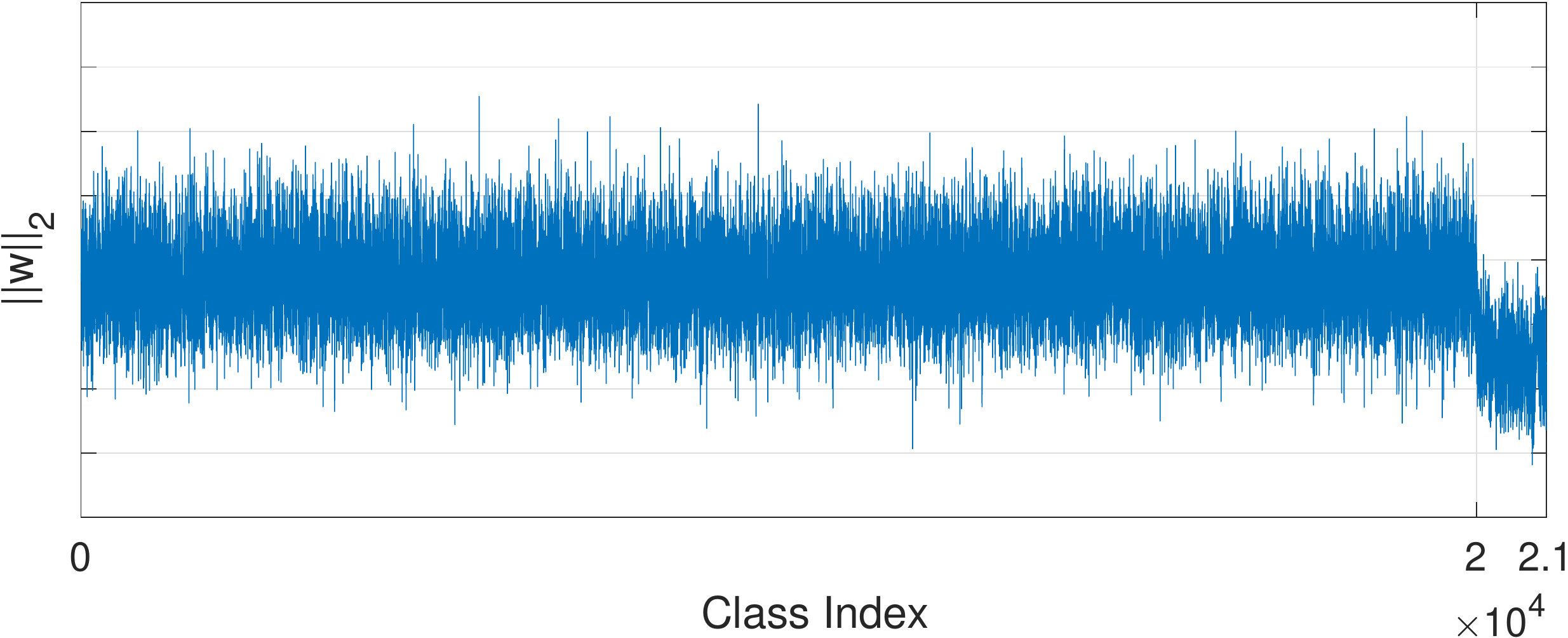}} \\[-0.05cm]
\caption{Norm of the weight vector $\mathbf{w}$ with standard MLR.  
The x-axis is the class index. 
The rightmost $1000$ classes on the x-axis correspond to the 
persons in the low-shot set. 
As shown in the figure,
without the UP term, $\|\mathbf{w}_k\|_2$ for the low-shot set 
is much smaller than that of the base set. 
}
\label{fig:wnorm1}
\end{figure}

\subsubsection{Underrepresented Classes Promotion}

In this subsection, 
we propose a method 
to promote the underrepresented classes, a.k.a. 
the classes with limited number of (or only one) samples. 
Our method is based on a prior 
which we design to increase the 
volumes of the partitions 
corresponding to the low-shot classes
in the feature space. 

Based on the previous analysis, 
we introduce a new term 
to the loss function
with the assumption 
that 
on average, the persons in the low-shot set 
and the persons in the base set
should have similar volume sizes 
for their corresponding partitions 
in the feature space. 
\begin{equation}
\label{eq:enorm}
\Lagr_{up} = 
\Lagr_s 
+   \left\| 
\frac{1}{|C_{l}|}
\sum_{k \in C_{l}}\|\mathbf{w}_k\|_2^2 - \alpha \right\|_2^2 \, , 
\end{equation}
where $\alpha$ is the parameter learned from the average of the squared norms of weight
vectors for the base classes, 
\begin{equation}
 \alpha \leftarrow \frac{1}{|C_{b}|} \sum_{k \in C_{b}} \| \mathbf{w}_k \|_2 ^2 .  
\end{equation}
We use $C_{b}$ and $C_{l}$ to denote the sets of the class indices for the \textit{base set} and the \textit{low-shot set}, respectively. 
As shown in Eq. \ref{eq:enorm}, the average of the squared norms of the weight vectors in the \textit{low-shot set} is promoted to the average of the squared norms of the weight vectors for the \textit{base set}. 
We call this term 
underrepresented-classes promotion (\textbf{UP}) term. 

For every mini-batch, we jointly optimize the cross entropy term and the UP
loss term. 
The derivative we sent back for back propagation 
is the summation of the 
derivative of cross entropy 
and the derivative of the UP term. 
We keep the rest of the optimization the same as a regular deep convolutional neural network. 

\subsubsection{Alternative Methods}
Adding extra terms of $\mathbf{w}_k$
to the cost function is essentially to inject prior knowledge to the system. 
Different assumptions
yield to different prior terms to the weight vectors. 
Here we discuss several alternatives for 
the UP-prior.

One typical method to handle insufficient data problem for regression and classification problems 
is to shrink $\mathbf{w}_k$,
\cite{Ti96,YandongLasso}. 
Here we choose the $L2$-norm option for optimization efficiency. 
\begin{equation}
\label{eq:snorm}
\Lagr_{l2} = 
\Lagr_s
+\sum_{k}  \|\mathbf{w}_k\|_2^2  \,. 
\end{equation}

Another option is to encourage all the weight vectors to have similar or even the same norms. 
A similar idea has been proposed in \cite{EWeightNorm} for the purpose of accelerating the training speed. 
We adopt the soft constraint on the squared norm of $\mathbf{w}$ here. 
\begin{equation}
\label{eq:fnorm}
\Lagr_{eq} = 
\Lagr_s
+ \sum_{k \in \{ C_{l} \cup C_{b} \} } \left\| \|\mathbf{w}_k\|_2^2 - \beta \right\|_2^2  \, , 
\end{equation}
where 
\begin{equation}
 \beta \leftarrow \frac{1}{|\{ C_{l}\cup C_{b} \}|} \sum_{k \in \{ C_{l}\cup C_{b} \}} \| \mathbf{w}_k \|_2 ^2 . \end{equation}
Note the major difference between this cost function and the cost function in Eq. \ref{eq:enorm} is that, 
in Eq. \ref{eq:fnorm}, the values of the norms of all 
$\mathbf{w}_k$ get affected and pushed to the same value, 
while in Eq. \ref{eq:enorm}, only the values of the norms of 
$\mathbf{w}_k$ for \textit{low-shot set} classes get promoted. 
The performance of all these options is presented in Section \ref{sec:experiments}. 

\section{Experimental Results}
\label{sec:experiments}

In this section, we list the experimental results 
for both the face representation model learning 
and multi-class classifier training. 

\subsection{Face Representation Learning}
\label{sec:experiments-feature}
Good feature representation model is the foundation of our task. 
In order to evaluate the discrimination and generalization capability of our face representation model, 
we leverage the LFW \cite{LFWTech,LFWTechUpdate}
verification task, 
which is to verify 
whether a given face pair (in total 6000)
belongs to the same person or not. 

We trained our face representation model 
using the images in our base set (already published to facilitate the research in the area, excluding people in LFW by design)
with ResNet-34 \cite{Resnet}. 
The verification accuracy with different models are listed in Table \ref{table:lfwResults}.
As shown, for the loss function, 
we investigated the standard cross entropy, 
cross entropy plus our CCS-loss term in Eq. \ref{eq:cca}, 
the center loss in \cite{centerECCV}, 
and the sphere face loss in \cite{sphereface}. 
For the CCS-loss, we set $\lambda$ in Eq. \ref{eq:cca} equal to $0.1$. 
For the center loss, we tried different sets of parameters and found the best performance could be achieved when the balancing coefficient was $0.005$, as reported in the table. 
For the sphere face \cite{sphereface},
we noticed this paper very recently and only tried 
limited sets of parameters  (there are four parameters to be adjusted together). 
The parameters reported in the paper can not make the network converge on our dataset. 
The only parameter set we found to make the network converge leads to worse results, compared with the standard cross-entropy loss. 

As shown in Table \ref{table:lfwResults}, 
we obtain the face representation model with the cutting-edge performance 
with the help of our CCS-loss term in 
Eq. \ref{eq:cca}. 
We follow the no-external data protocol and use the CCS model to investigate the one-shot learning phase. 

For comparison, we also list
the results from other methods referring the numbers stated in the published corresponding papers. 
These methods use different datasets and different networks structures. 
Please note that applying our CCS face with the public available MS-Celeb-1M dataset
leads to better performance on LFW verification task
compared with the other methods with public/private datasets, which demonstrates the effectiveness of our CCS method.

\begin{table}
\begin{center} 
\begin{tabular}{|l||c|c|}
\hline
Methods &Network&Accuracy \\
\hline\hline
Cross entropy only  & 1 & $98.88\%$  \\
\hline
CCS face in \ref{eq:cca} (ours)  & 1 & $\mathbf{99.28}\%$  \\
\hline
Center face \cite{centerECCV}  & 1 & $99.06\%$  \\
\hline
Sphere face \cite{sphereface} & 1 & $-.--\%$  \\
\hline
\end{tabular}
\end{center}
\caption{
LFW verification results obtained with models trained with 
the same dataset (base set). 
All the models use ResNet-34 \cite{Resnet} as the feature extractor
to highlight the effectiveness of the loss function design. 
For the sphere face, we do not find a set of parameters which works for this dataset, with limited number of trials.
}
\label{table:lfwResults}
\end{table}

\begin{table}
\begin{center} 
\begin{tabular}{|l||c|c|c|}
\hline
Methods &Dataset&Network&Accuracy \\
\hline\hline
JB \cite{JB}& Public & -- & $96.33\%$  \\
\hline
Human & -- & -- & $97.53\%$ \\
\hline
DeepFace\cite{vgg_face}  & Public & 1 & $97.27\%$ \\
\hline
DeepID2,3 \cite{Xiaoou_Deep2,Xiaoou_Deep3}& Public & 200 & $99.53\%$  \\
\hline
FaceNet \cite{Google_Face} & Private & 1 & \textcolor{red}{$99.63\%$} \\
\hline
Center face \cite{centerECCV}  & Private & 1 & $99.28\%$ \\
\hline
Center face \cite{sphereface}  & Public & 1 & $99.05\%$  \\
\hline
Sphere face \cite{sphereface} & Public & 1 & $99.42\%$  \\
\hline
CCS face (ours) & Public & 1 & \textcolor{blue}{$\mathbf{99.71}\%$}  \\
\hline
\end{tabular}
\end{center}
\caption{
For reference, 
LFW verification
results reported in the peer-reviewed publications (partial list). 
Different datasets and network structures were used.
For CCS-face, we used ResNet-34 and a cleaned version of the full MS-Celeb-1M data.
The \textit{closest runner-up} to our CCS-face 
is 
FaceNet in \cite{Google_Face}, 
which is trained with 
$>100$M \textbf{private} images for $8$M persons, while
our model is reproducible.}
\label{table:lfwResults-r}
\end{table}

We also tried different values of $\lambda$ in Eq. \ref{eq:cca} and found our method is not sensitive to the choose of $\lambda$, shown in the Table \ref{table:lambda}. 
Larger $\lambda$ means stronger regularizer applied. 
Note $\lambda = 0$ corresponds to no CCS-loss applied. 
\begin{table}[h]
\begin{center}
\begin{tabular}{|l||c|c|c|c|c|}
\hline
$\lambda$ & $0$ &$0.01$&$0.1$ & $1$ &$10$ \\
\hline\hline
LFW & $98.88\%$ & $90.05\%$  & $99.28\%$ &$99.20\%$ & $99.20$ \\
\hline
\end{tabular}
\end{center}
\caption{
LFW verification results obtained with 
different $\lambda$ 
in Eq. \ref{eq:cca}. 
Larger $\lambda$ means stronger regularizer applied.
}
\label{table:lambda}
\end{table}

\subsection{One-shot Face Recognition}

In phase two, 
we train a $21$K-class classifier 
to recognize the persons in both the base set and the low-shot set. 
Since there is only one image per person for training in the low-shot set, 
we repeat each sample in the low-shot set for $100$ times through all the experiments in this section. 
In order to test the performance, we apply this classifier 
with $120,000$ test images consists of images from the base or low-shot set. 
We focus on the recognition performance 
in the novel set 
while monitoring the recognition performance in the base set
to ensure that the performance improvement in the novel set
does not harm the performance in the base set. 

To recognize the test images for the persons in the novel set is a challenging task. 
The one training image per person was randomly preselected, 
and the selected image set
includes images of low resolution, 
profile faces, and faces with occlusions. 
We provide more examples in the supplementary materials due to space constraint. 
The training images in the novel set 
show a large range of variations
in 
gender, race, ethnicity, age, 
camera quality (or evening drawings),
lighting, focus, pose, expressions, and many other parameters. 
Moreover, we applied de-duplication algorithms to ensure that the training image is visually different from the test images, 
and the test images can cover many different looks for a given person.

\begin{table}
\begin{center}
\begin{tabular}{|l|c|c|}
\hline
Method & C@$99\%$ & C@$99.9\%$  \\
\hline\hline
Fixed Feature & $25.65\%$ & $0.89\%$\\
\hline
SGM \cite{Ross2016lowshot} & $27.23\%$ &$4.24\%$ \\
\hline
Update Feature & $26.09\%$ & $0.97\%$\\
\hline
Direct Train & $15.25\%$ & $0.84\%$\\
\hline
Shrink Norm (Eq.\ref{eq:snorm}) & $32.58\%$ &$2.11\%$\\
\hline
Equal Norm (Eq.\ref{eq:fnorm}) & $32.56\%$ &$5.18\%$\\
\hline
UP Only (Eq.\ref{eq:enorm}) &${77.48}{\%}$ & ${47.53}{\%}$  \\
\hline
CCS Only (Eq.\ref{eq:cca}) &${62.55}{\%}$ & ${11.13}{\%}$  \\
\hline
\textbf{Our:} CCS (\ref{eq:cca}) plus UP (\ref{eq:enorm}) &$\mathbf{94.89}\mathbold{\%}$ & $\mathbf{83.60}\mathbold{\%}$  \\
\hline
\hline
Hybrid \cite{ICCV-W-Yue} & $92.64\%$ & N/A \\
\hline
Doppelganger \cite{ICCV-W-Doppelganger} & $73.86\%$ & N/A  \\
\hline
Generation-based \cite{ICCV-W-Generation} & $61.21\%$ & N/A \\
\hline
\end{tabular}
\end{center}
\caption{Coverage at Precisions = $99\%$ and $99.9\%$ on the \textbf{low-shot set}. 
Please refer to subsection 5.2 or the corresponding citations for the detailed descriptions for all the methods. 
As shown in the table, our method CCS+UP 
loss significantly improves the recall at precision $99\%$ and $99.9\%$ 
and achieves the \textbf{best} performance among all the methods.
Unless specified with ``CCS'', or numbers reported by other papers 
(Hybrid, Doppelganger, and Generation-based)
\cite{ICCV-W-Yue,ICCV-W-Doppelganger,ICCV-W-Generation}, 
the face feature extractor was trained with cross entropy loss. 
}
\label{tab:core}
\end{table}

The methods we experimented and the corresponding results 
are listed in Table \ref{tab:core}. 
We also list the performance from the \textit{top-3} methods presented in
the MS-Celeb-1M challenge in 
ICCV 2017 workshop
for this task. 
We use coverage rate at precision $99\%$ and $99.9\%$ as our evaluation metrics
since this is the major requirement for a real recognizer. 
The methods in the table are described as follows. 


The ``Fixed Feature'' in Table \ref{tab:core}
means that,
in phase two, 
we do not update the feature extractor and only train the classifier 
in Eq. \ref{eq:crossentropy} 
with the feature extractor
in phase one. 

The SGM, known as squared gradient magnitude loss,
is obtained by updating the feature extractor during phase one using the feature shrinking method as described in \cite{Ross2016lowshot}. 
Compared with the ``Fixed-Feature'', 
SGM method introduces about $2\%$ gain in coverage 
when precision requirement is $99\%$, 
while $4\%$ gain when precision requirement is $99.9\%$. 
The improvement for face recognition by feature shrinking in \cite{Ross2016lowshot}
is not as significant as that for general image. 
The reason might be that
the face feature is already a good representation for faces
and the representation learning is not the main bottleneck. 
Note that we did not apply the feature hallucinating method as proposed in \cite{Ross2016lowshot}
for fair comparison and 
to highlight the contribution of model learning, 
rather than data augmentation. 
To couple the feature hallucinating method (may need to be modified for face) is a good direction for the next step. 

The ``Update Feature'' method in Table \ref{tab:core} means that
we fine-tune the feature extractor simultaneously when we train the classifier in Eq. \ref{eq:crossentropy}
in phase two.
The feature updating does not change 
the recognizer's performance
too much. 

The rest three methods (shrink norm, equal norm, UP-method)
in Table \ref{tab:core}
are obtained by using the cost functions defined in Eq. \ref{eq:snorm}, Eq. \ref{eq:fnorm}, and Eq. \ref{eq:enorm}
as supervision signals for deep convolutional neural network in phase two, 
with the face feature updating option.   
As shown in the table, our UP term improves the coverage@precision=$99\%$ and coverage@precision=$99.9\%$ significantly. 

The coverage at precision $99\%$ on the base set 
obtained by using any classifier-based methods in Table \ref{tab:core}
is $100\%$. The top-1 accuracy on the base set obtained by any of these classifier-based methods
is $99.80\pm0.02\%$. 
Thus we do not report them separately in the table.


\section{Conclusion and Future Work}
In this paper, we have studied the problem of one-shot face recognition. 
We build a solution for this task from two perspectives. 
First,
we introduce a method called 
Classification vector-centered Cosine Similarity (CCS) to train a better face feature extractor, which 
has better discrimination capability for persons with limited number of
training images, compared with the extractor trained without CCS. 
Second, 
we reveal that the deficiency of multinomial logistic regression in one-shot learning is related to the norms of the weight vectors in multinomial logistic regression, and propose a novel loss term called underrepresented-classes promotion to effectively address the data imbalance problem in the one-shot learning. 
The evaluation results on the benchmark dataset show that the two new loss terms together bring a significant gain by improving the recognition coverage rate from $25.65\%$ to $94.89\%$ at the precision of $99\%$ for one-shot classes, while still keep an overall accuracy of $99.8\%$ for normal classes.

In the future, we 
are interested in applying 
the UP prior and CCS, and exploring
ore options to improve low-shot learning in the general visual recognition problems.


{\small
\bibliographystyle{ieee}
\bibliography{face}
}
\section{Appendix}
We summarize 
some potential questions
and response as follows. 
Most of the information is included in the main paper, 
yet we 
re-organize and add more details 
using the QA structure
for the convenience of the readers.

\subsection{KNN vs. Softmax}
 
To leverage face verification (pairwise comparison) to solve the face identification problem is a straight-forward and popular solution \cite{UW_MegaFace}. We are also suggested to use k-nearest neighbors (KNN). Here we discuss why we don't choose KNN. 


\begin{figure}[h]
\centering
{\includegraphics[width=0.99\linewidth]{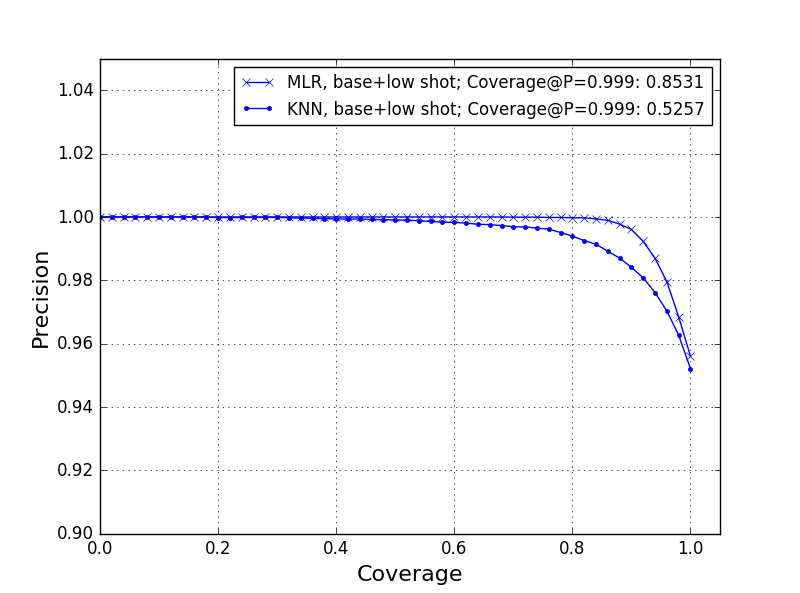}}
\caption{Deep comparison of K nearest-neighborhood (KNN)
and multinomial logistic regression (MLR, a.k.a, Softmax)
applied on top of the feature extraction for large-scale face identification.
The figure shows the precision and coverage 
on \textbf{all the test images from both the base and low-shot sets}.
The same feature extractor was used for both the method. 
As shown, 
though both the methods lead to similar Top-1 accuracy, 
MLR has much larger coverage at high precision compared with that of KNN (MLR: $85.31\%$ @ Precssion = $99.9\%$ while KNN: $52.57\%$ @ Precssion = $99.9\%$). 
The major reason is that MLR update the classification weight vector using information from 
the samples from both the corresponding class and other classes. 
}
\label{fig:MLR-KNN}
\end{figure}

We acknowledge the advantages of KNN (or other similar template-based method):
no classifier-training is needed after the feature representation model is trained, 
and KNN does not suffer much from imbalanced data.
If the feature extraction is perfect (the distance between samples from the same class is always smaller than the distance between samples from different classes), KNN is good enough for any face identification problem. 
However, there is no perfect face feature though a lot of progress has been made along this direction. 

Given a reasonably good, yet not perfect face feature extractor, 
our experimental results demonstrate that 
with the large-scale face identification setup, 
the multinomial logistic regression (MLR, a.k.a. Softmax) has better performance than that of KNN. 
As shown in Fig. \ref{fig:MLR-KNN}, 
both KNN and MLR (with the same feature extractor for fair comparison) were tested with all the test images from both the base and low-shot set. 
Though they lead to similar Top-1 accuracy, 
MLR has much larger coverage at high precision compared with that of KNN (MLR: $85.31\%$ @ Precssion = $99.9\%$ while KNN: $52.57\%$ @ Precssion = $99.9\%$).

In the previous publication  
\cite{ICCV-W-Yue,ACMMMMSCeleb1M-1,ACMMMMSCeleb1M-2,Xu_2017_ICCV} on large-scale face recognition, 
authors have made similar statement. 
We believe the major reason is that
in MLR, the weight vectors for each of the classes
are estimated using discriminant information from all the classes, 
while in the KNN setup, 
the query image only needs to be close enough to one local class 
to be recognized.

Moreover, for the KNN method, 
if we use all the face images for every person in the gallery, 
the complexity is usually too high for large scale recognition,
and the gallery dataset needs to be very clean to ensure the high precision. 
If we do not keep all the images per person, 
how to construct representer for each class is still an open problem. 
On the contrary,
with MLR, after the feature extraction,  
the computational complexity of estimating the persons' identity 
is linear to the number of persons, 
not the number of images in the gallery.

\subsubsection{Challenge of Imbalanced Training data}
\begin{figure}[h]
\centering
{\includegraphics[width=0.99\linewidth]{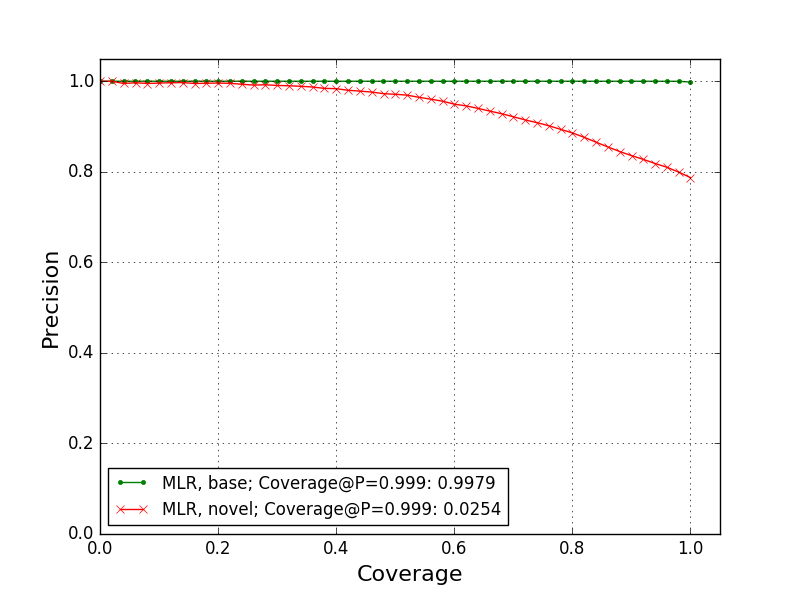}}
\caption{Precision and coverage of MLR (a.k.a, Softmax)
evaluated on the base and low-shot sets separately. 
As shown, 
Though MLR has overall better performance over KNN, as shown in Fig. \ref{fig:MLR-KNN}, 
the standard MLR classifier 
suffers from the imbalanced training data
and has poor performance 
with the low-shot classes
even these classes are oversampled during training.  
}
\label{fig:MLR}
\end{figure}

Though MLR has overall better performance over KNN, as shown in Fig. \ref{fig:MLR-KNN}, 
the standard MLR classifier 
suffers from the imbalanced training data
and has poor performance 
with the low-shot classes
even these classes are oversampled during training. 
We evaluate the precision and recall of MLR on the base and low-shot sets separately and present the results in Fig. \ref{fig:MLR}. 
As shown, 
though MLR has overall better performance over KNN, 
the standard MLR classifier 
suffers from the imbalanced training data
and has poor performance 
with the low-shot classes
even these classes are oversampled during training.  

Recently, some works develop hybrid solutions by combining MLR and KNN \cite{ICCV-W-Yue,Xu_2017_ICCV} and achieve promising results. 
In these work, when MLR does not have high confidence (threshold tuning is needed), 
KNN is used. 

We solve the training data imbalance challenge from a different perspective.
Different from the hybrid solution, our solution only has one MLR as the classifier so that no threshold is needed to switch between classifiers. 
We boost the performance of MLR by regularizing the norm of the weight vectors in MLR. We have not seen a lot of effort in this direction, especially in the deep learning scenario. 

\subsection{Feature Learning}
There have been many efforts in improving the face feature learning by adding regularizers to the Softmax loss term. 
Maybe the early pioneers in this direction are the center-loss in \cite{centerECCV} or the similar version called dense loss in \cite{Latha:dense}. 

We find that better face feature model always help the final results, no matter KNN, or MLR, or MLR with underrepresented class promotion method is used. Therefore, we also investigate how to learn an even better face feature extractor.  

Our proposed classification vector-centered cosine similarity (CCS) term is different from its cousin
center face \cite{centerECCV} or sphere face \cite{sphereface}. 
We try to minimize the angle between the feature vectors and the corresponding weights
while
center-loss minimizes the distance between the feature vectors
and the corresponding class centers.  
SphereFace 
emphasizes on 
maximizing the margin (angle) between the features and the corresponding decision boundary. 

Experimental results show that 
our method has better performance for our task,
when the same training data is used, as listed in the main paper. 
SphereFace has four task-specific hyper-parameters to control the regularization strength. 
Given the limited time (we noticed this work short time before submission), 
we failed to find good hyper-parameters for our task (also tried authors' parameters). 
 
Please also note that the improvement on the
face representation model is one of the three contributions of our paper (the other two are the benchmark task design and UP-term for data imbalance).

\subsection{Practical Applications}
In this paper, we study the problem of
training a large-scale face identification model
using \textit{imbalanced} training images for a large quantity of persons, 
and then use this model to identify other face images
for the persons in \textit{the same group}. Though this setup may not be the case for video surveillance where the person to be recognized is not typically in the training data, 
this setup is still widely used 
when the images for the persons to be recognized are available beforehand, 
and an accurate recognizer is needed for a large
and relatively fixed group of persons. 
For example, 
large-scale celebrity recognition for search engine, 
public figure recognition for media industry, 
and movie character annotation for video streaming companies. This one-shot face recognition challenge has been 
selected as one of the ICCV 2017 workshops, and 
attracted 
more than $40$ teams registered last year and hundreds times of data download. 

\subsection{More discussion on UP term}
For the reading convenience, we include the figure to demonstrate the impact of underrepresented class promotion (UP) loss term on the norm of $\mathbf{w}$ here. 
\begin{figure}[h!]
\centering
\subfloat[Without UP]
{\includegraphics[width=0.8\linewidth]{figs/wnorm-noalignb-iter100000.pdf}} \\
\subfloat[With UP]
{\includegraphics[width=0.8\linewidth]{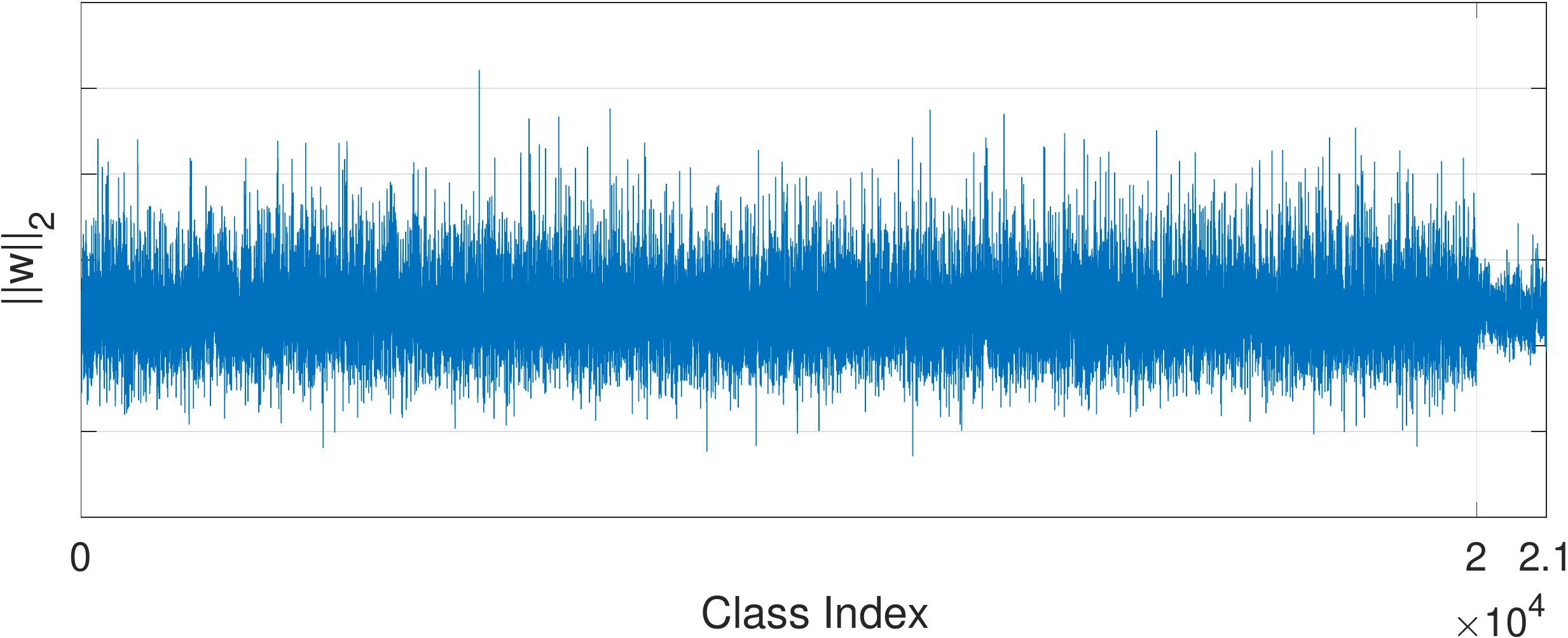}} \\[-0.05cm]
\caption{Norm of the weight vector $\mathbf{w}$ with standard MLR with/without UP.  
The x-axis is the class index. 
The rightmost $1000$ classes on the x-axis correspond to the 
persons in the low-shot set. 
As shown in the sub-figure [a],
without the UP term, $\|\mathbf{w}_k\|_2$ for the low-shot set 
is much smaller than that of the base set, 
while with the UP term, on average, 
$\|\mathbf{w}_k\|_2$ for the low-shot set tends 
to have similar values as that of the base set.
}
\label{fig:wnorm1}
\end{figure}

As shown in the sub-figure [a],
without the UP term, $\|\mathbf{w}_k\|_2$ for the low-shot set 
is much smaller than that of the base set, 
while with the UP term, on average, 
$\|\mathbf{w}_k\|_2$ for the low-shot set tends 
to have similar values as that of the base set.
Note that 
the variance of the norm of $\mathbf{w}$ for different low-shot classes is reduced.
This is a byproduct of the UP-term: since all the $||\mathbf{w}_k||$ for the low-shot classes are encouraged to be closer to one scalar value $\alpha$, they naturally have similar values (not identical though), which leads to smaller variance of $\mathbf{w}_k$ among the low-shot classes. The UP-term does not change $w_k$ for base classes. How to more actively control the variance is still an open problem. We will study in the future.



\end{document}